\documentclass{article}
\usepackage{spconf,amsmath,graphicx,hyperref}

\usepackage[table]{xcolor}
\usepackage{subcaption}
\usepackage{amssymb}
\usepackage{booktabs}

\newcommand{\method}{RDT-3D}
\newcommand{\rdt}{RDT}
\newcommand{\dataset}{Dual3D-300K}
\newcommand{\Rg}{R_{\mathrm{g}}}   
\newcommand{\Rs}{R_{\mathrm{s}}}   
\newcommand{\bx}{\mathbf{x}}
\newcommand{\bp}{\mathbf{p}}
\newcommand{\bz}{\mathbf{z}}

\title{RDT-3D: Resolution-Decoupled Training for High-Fidelity 3D Geometry Generation}

\name{Jiale Zhang, Yang Liu, Chengyang Li, Jin Yuan, Qianxi Jia, Taicheng Huang\textsuperscript{*}
\thanks{\textsuperscript{*}Corresponding author: huangtaicheng.1@jd.com}}
\address{JD.com, Beijing, China}

\begin{document}
\maketitle

\begin{abstract}
High-fidelity 3D geometry generation requires maintaining global structural coherence alongside fine-grained surface details. While recent coarse-to-fine frameworks leverage Rotary Position Embeddings (RoPE) to condition local queries, we reveal a critical training-inference resolution discrepancy: coordinate guidance and geometric supervision are aligned during training, whereas inference decodes high-resolution geometry from coarse spatial coordinates. This mismatch introduces a severe distribution shift that degrades fine details. To overcome this, we propose \textit{Resolution-Decoupled Training} (\rdt{}), which anchors RoPE guidance at $512$ resolution while supervising geometry at $1536$, enabling robust detail extrapolation. Additionally, to eliminate high-resolution data bottlenecks, we present \dataset{}, a large-scale paired dataset comprising $300\text{K}$ watertight 3D point clouds explicitly aligned across $512$ and $1536$ resolutions. Extensive experiments demonstrate that \rdt{} substantially improves sharp geometric regions, while \dataset{} offers a scalable benchmark for high-resolution 3D generation.

\end{abstract}


\vspace{-1em}
\section{Introduction}
\label{sec:intro}

High-fidelity 3D assets are fundamental to gaming, film, and immersive applications. Driven by generative AI, 3D content creation has advanced rapidly~\cite{kerbl20233d,zhang2024clay,hunyuan3d2025hunyuan3d,xiang2025structured,xiao2026sparse,li2026sparc3d}. However, synthesizing complex 3D geometry requires balancing global structural coherence with fine-grained local details. Recent coarse-to-fine paradigms tackle this via a two-stage pipeline: a coarse model establishes global topology, followed by a refinement stage that predicts localized surface details. State-of-the-art methods, such as UltraShape~\cite{jia2025ultrashape} and Lattice~\cite{lai2026lattice}, condition localized voxel queries using Rotary Position Embeddings (RoPE)~\cite{su2024roformer} for precise spatial synthesis.

Despite their success, we reveal a critical yet overlooked bottleneck: a resolution discrepancy between training and inference. Existing models are trained symmetrically, where coordinate guidance and geometric supervision operate at identical spatial resolutions. During inference, however, the model functions asymmetrically—decoding high-resolution geometry ($1536$) from coarse positional guidance ($512$). This mismatch introduces a severe distribution shift, preventing the refinement network from robustly extrapolating high-frequency surface details from coarse spatial cues.

To overcome this limitation, we propose \textit{Resolution-Decoupled Training} (\rdt{}), a novel framework designed for high-resolution 3D refinement. \rdt{} explicitly decouples spatial localization from geometric reconstruction during training. Specifically, we anchor the RoPE guidance at a coarse resolution ($512$) to replicate inference conditions, while enforcing geometric supervision at ultra-high resolutions ($1536$). This asymmetric design forces the network to learn robust mapping from coarse spatial inputs to fine-grained ground-truth surfaces.

Additionally, to address the infrastructure bottleneck of processing high-resolution watertight meshes~\cite{deitke2023objaverse,hunyuan3d2025hunyuan3d}, we introduce and will open-source \dataset{}, a large-scale paired dataset comprising $300\text{K}$ 3D point clouds aligned at both $512$ and $1536$ resolutions. 

Our main contributions are summarized as follows:
\begin{itemize}
    \vspace{-0.5em}\item We identify a key resolution mismatch in RoPE-guided 3D generation and propose \textit{Resolution-Decoupled Training} (\rdt{}), an asymmetric learning scheme that resolves training-inference inconsistency.
    
    \vspace{-0.5em}\item We release \dataset{}, a paired dataset of $300\text{K}$ watertight 3D point clouds explicitly aligned across $512$ and $1536$ resolutions, lowering the computational barrier for high-resolution 3D generation.
    
    \vspace{-0.5em}\item Comprehensive evaluations demonstrate that \rdt{} significantly outperforms state-of-the-art baselines, producing vastly superior geometry with sharp edges, thin structures, and intricate surface topologies.
\end{itemize}

\begin{figure*}[t]
    \centering
    \includegraphics[width=\linewidth]{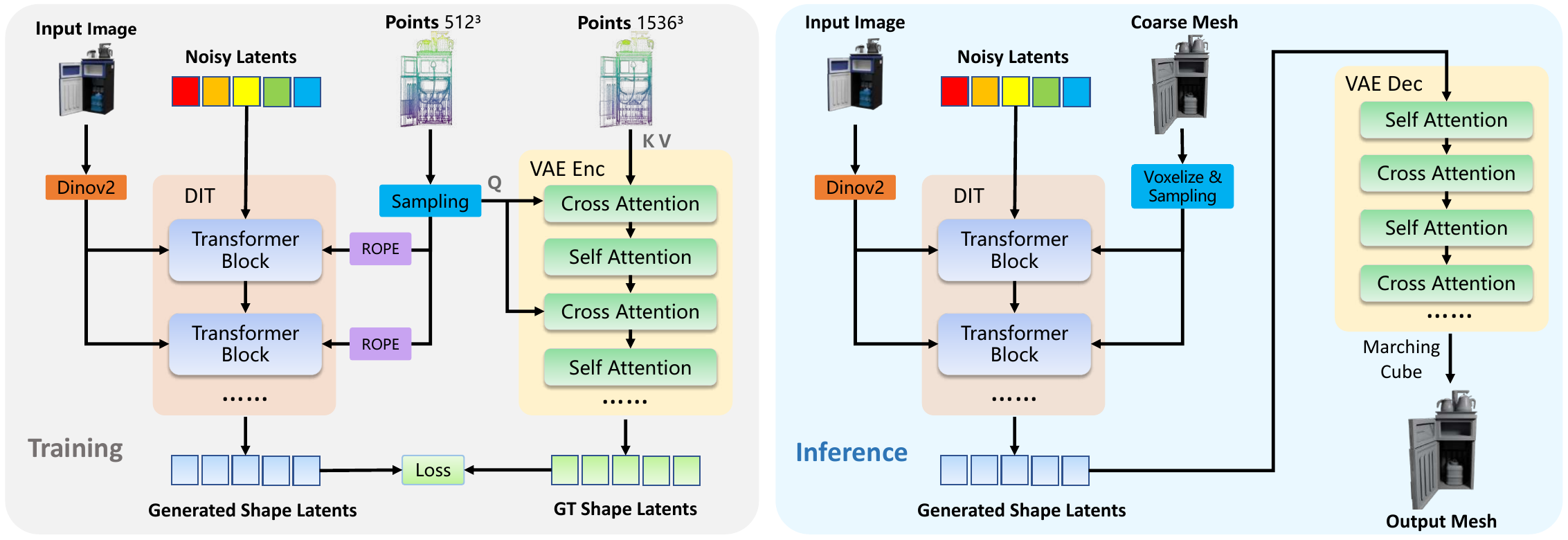}
    \caption{\textbf{Overview of the proposed RDT-3D framework.} 
    \textbf{Training (Left):} RDT-3D decouples spatial guidance from geometric supervision. Coarse point clouds ($512^3$) provide spatial query locations injected into the DiT via ROPE, while fine point clouds ($1536^3$) are encoded into high-resolution GT shape latents for diffusion training. 
    \textbf{Inference (Right):} Spatial queries sampled from a coarse mesh guide the trained DiT to generate fine shape latents, which are subsequently decoded into a high-fidelity 3D mesh via Marching Cubes.}
    \label{fig:pipeline}
    \vspace{-1em}
    \label{pipeline}
\end{figure*}

\begin{figure}[t]
  \centering
  \includegraphics[width=\linewidth]{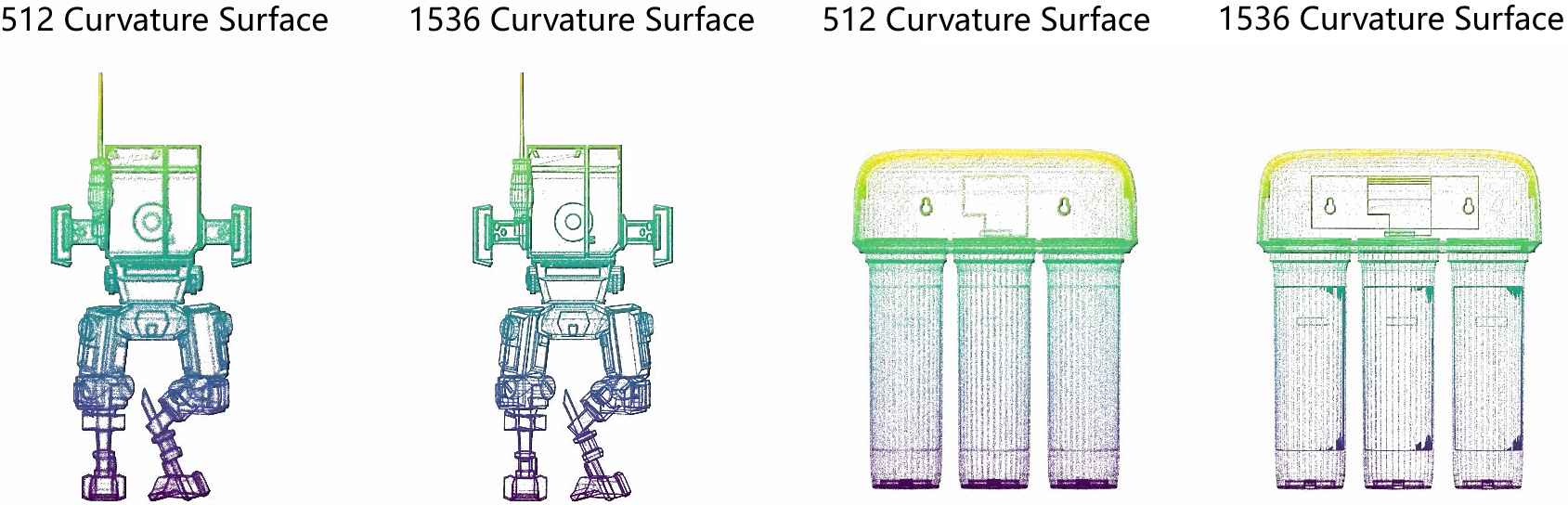}
  \caption{\textbf{Visual cases in \dataset{}.} Higher resolution ($1536$) captures significantly finer geometric details and sharper surfacecompared to the base resolution ($512$). \textit{(Best viewed when zoomed in.)}}
  \vspace{-1em}
  \label{fig:dataset_samples}
\end{figure}


    
    

\vspace{-1.0em}
\section{Method}
\vspace{-0.5em}
\label{sec:method}

We introduce \method{}, an end-to-end framework for high-fidelity 3D geometry generation (Fig.~\ref{fig:pipeline}). Unlike prior methods~\cite{lai2026lattice,jia2025ultrashape} that couple spatial guidance with geometric supervision at identical resolutions, \method{} explicitly decouples them during training. Specifically, it anchors RoPE coordinate guidance at a coarse resolution to match inference conditions, while enforcing target latent supervision at an ultra-high resolution. During inference, given a single image, the framework first obtains a coarse mesh via Hunyuan3D-2.1~\cite{hunyuan3d2025hunyuan3d} or TRELLIS~\cite{xiang2025structured}, extracts spatial queries via voxelization and point sampling for RoPE guidance, and finally decodes the refined latents into a high-fidelity 3D mesh.




\vspace{-1em}
\subsection{The Resolution Discrepancy}
\label{sec:gap}

Let $\Rg$ denote the spatial resolution used to construct the RoPE coordinate guidance $P$, and $\Rs$ denote the resolution at which target geometry is supervised and decoded (i.e., the spatial fidelity of target latents $\bz_0$ and the Marching Cubes mesh extraction).

\noindent\textbf{Conventional training} derives both guidance and supervision from an identical high-resolution ground truth, enforcing $\Rg=\Rs$. Consequently, the model exclusively observes relative positional offsets $\bp_j-\bp_i$ evaluated on the fine grid $R$.

\noindent\textbf{Inference}, however, is inherently asymmetric: positional guidance is derived from a coarse spatial proxy (e.g., $\Rg=512$, sampled from a stage-1 coarse mesh), whereas target geometry must be decoded at an ultra-high resolution (e.g., $\Rs=1536$). This induces an intrinsic resolution gap: $\Rg {<} \Rs$.

This mismatch induces a severe distribution shift in the RoPE coordinate signals between training and inference. During conventional training, spatial anchors are sampled from the accurate ground-truth geometry, yielding precise positional cues. At inference, however, spatial anchors are derived from the stage-1 coarse mesh, which inherently contains surface localization errors and missing details. Having been trained exclusively on precise positional priors, the refinement network struggles to disambiguate fine geometric details under imprecise positional guidance—manifesting as oversmoothed thin structures and rounded sharp edges.

\begin{figure*}[t]
  \centering
  \includegraphics[width=\linewidth]{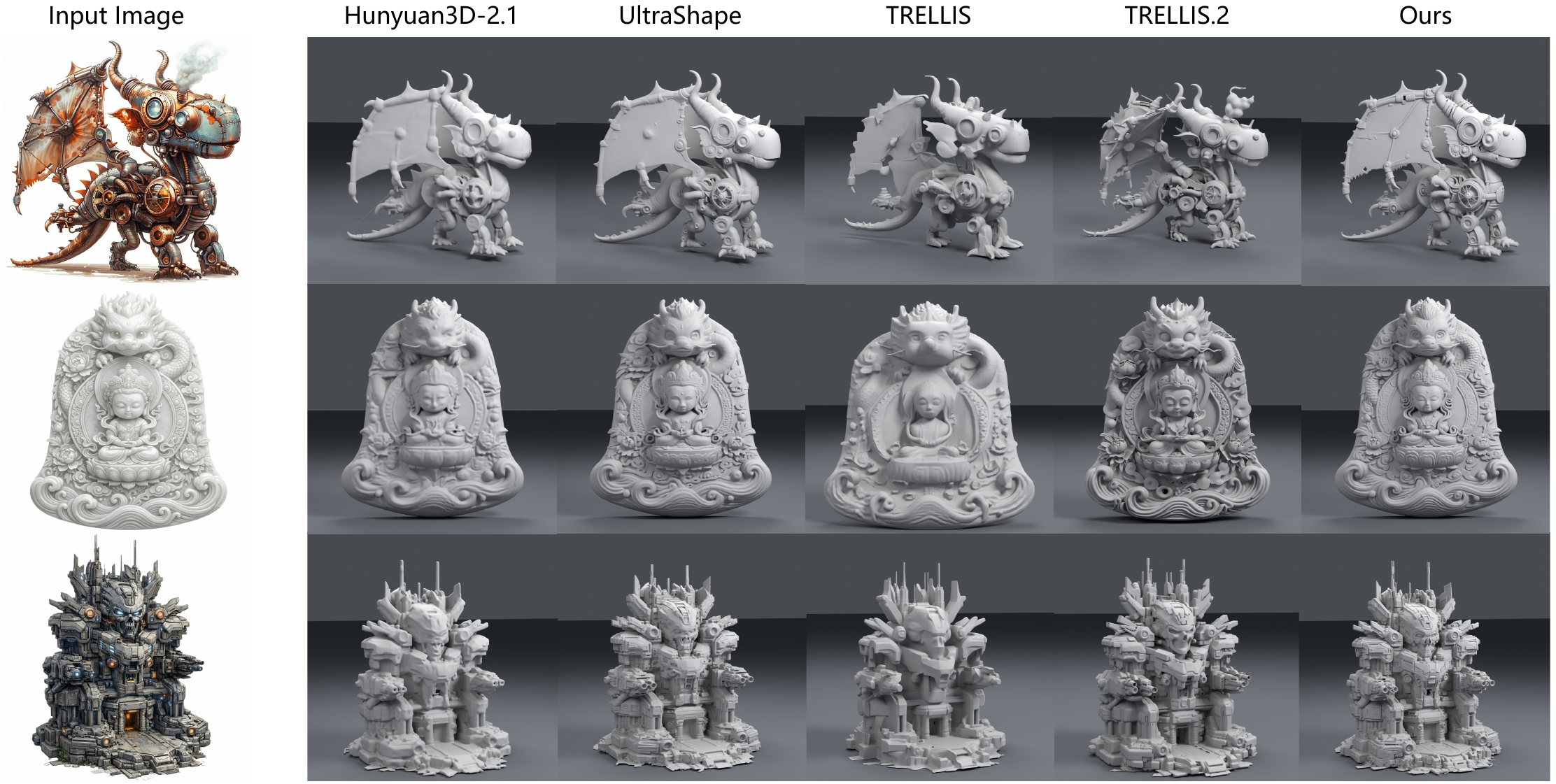}
  \caption{\textbf{Qualitative comparison with state-of-the-art methods.} 
  Compared to existing approaches, our method captures significantly finer geometric details and produces sharper surface reconstructions.}
  \vspace{-0.5em}
  \label{fig:qualitative_comparison}
\end{figure*}

\begin{table}[t]
\centering
\caption{\dataset{} Sampling \& Property Statistics.}
\label{tab:sampling_stats}
\resizebox{\linewidth}{!}{%
\begin{tabular}{lccccc}
\toprule
Sampling Type & Category & Res. (coarse/fine) & \#Pts/sample & Normals \\
\midrule
Uniform Surface   & Surface Points & $512\,/\,1536$ & 300K & \checkmark \\
Curvature Surface & Surface Points & $512\,/\,1536$ & 300K & \checkmark \\
Uniform Near      & Near SDF       & $512\,/\,1536$ & 600K & \checkmark \\
Curvature Near    & Near SDF       & $512\,/\,1536$ & 300K & \checkmark \\
Free-Space        & Space SDF      & $512\,/\,1536$ & 400K & --         \\
\midrule
\textbf{Total}    & --             & --             & \textbf{1.9M} & \checkmark \\
\bottomrule
\vspace{-2.5em}
\end{tabular}}
\end{table}

\vspace{-1em}
\subsection{Resolution-Decoupled Training}
\label{sec:rdt}

To eliminate the discrepancy identified in Section~\ref{sec:gap}, we introduce \emph{Resolution-Decoupled Training} (\rdt{}). Instead of coupling guidance and supervision at the same resolution, \rdt{} explicit decouples the RoPE coordinate guidance resolution $\Rg$ from the target supervision resolution $\Rs$. By training on coarse spatial coordinates alongside high-resolution geometric ground truth, the refiner is compelled to extrapolate fine-grained surfaces from imprecise spatial cues.

\noindent\textbf{Coarse Localization Branch.}
To mirror inference-time spatial cues, RoPE guidance is constructed by quantizing anchor locations $\bx_i$ onto a coarse grid ($\Rg{=}512$). Because spatial coordinates derived from lower-resolution sampling naturally exhibit localization noise and lack fine-grained structural alignment, the network is restricted from over-relying on perfect positional priors, forcing it to infer precise geometry from visual conditions and contextual features.

\noindent\textbf{Fine Supervision Branch.}
Concurrently, target latents $\bz_0$ are encoded from a high-resolution surface $S_{\Rs}$ ($\Rs{=}1536$), preserving sharp edges and thin structures. Formally, the paired training configuration is:
\vspace{-0.25em}
\begin{equation}
\begin{aligned}
\text{Coarse Guidance:} \quad & \bp_i = \phi_{\Rg}(\bx_i) \quad (\Rg{=}512), \\
\text{Fine Supervision:} \quad & \bz_0 = \mathcal{E}(S_{\Rs}) \quad (\Rs{=}1536).
\end{aligned}
\label{eq:rdt}
\end{equation}
\vspace{-0.25em}
Because $\Rg{<}\Rs$, latents sharing identical coarse positional cues are supervised toward \emph{distinct} high-resolution geometric targets. This training regime conditions $v_\theta$ to robustly recover high-frequency detail under approximate positional guidance—aligning directly with test-time conditions.

\noindent\textbf{Guidance Rationale.} 
Directly using coarse meshes generated by upstream models (e.g., Hunyuan3D-2.1) for training introduces excessive noise and severe structural divergence from ground-truth targets, causing severe training instability. Instead, utilizing low-resolution ground-truth meshes maintains global structural alignment with target supervision while naturally modeling the coarse-to-fine mapping, ensuring both robust feature learning and stable optimization.

\vspace{-1em}
\subsection{Loss Function}
\vspace{-0.5em}
\label{sec:loss}

We train $v_\theta$ using a rectified-flow objective~\cite{lipman2022flow,esser2024scaling}. For target latents $\bz_0 = \mathcal{E}(S_{\Rs})$ and noise $\epsilon \sim \mathcal{N}(0, \mathbf{I})$, we interpolate $\bz_t = (1-t)\bz_0 + t\epsilon$ ($t \sim \mathcal{U}(0,1)$) with velocity $u_t = \epsilon - \bz_0$. The loss is:
\vspace{-0.25em}
\begin{equation}
\mathcal{L}_{\text{RDT}} = \mathbb{E}_{t, \bz_0, \epsilon} \Big[ \big\| v_\theta(\bz_t, t, c, P) - (\epsilon - \bz_0) \big\|_2^2 \Big],
\label{eq:fm}
\end{equation}
\vspace{-0.25em}
where $c$ is the image condition and $P = \{\phi_{\Rg}(\bx_i)\}_{i=1}^N$ represents coarse RoPE guidance. 

\vspace{-0.5em}
\section{The \dataset{} Dataset}
\vspace{-0.5em}
\label{sec:data}
To enable high-resolution 3D geometry generation, we introduce \dataset{}, a benchmark of $300\text{K}$ pre-aligned watertight point cloud pairs ($512$ coarse for RoPE guidance and $1536$ fine for geometric supervision) and rendered images . Compared to HY3D-Bench~\cite{hunyuan3d2026hy3d} ($250\text{K}$ samples, resolution $\le 512$), \dataset{} scales up dataset size and elevates geometric resolution to $1536$.


\noindent\textbf{Data curation \& canonicalization.}
\dataset{} is curated from diverse sources, including our in-house high-quality assets, ABO~\cite{collins2022abo}, 3D-FUTURE~\cite{fu20213d}, Objaverse(XL)~\cite{deitke2023objaverse}, and TexVerse~\cite{zhang2025texverse}. We retain manifold meshes while filtering out low-face geometries. Each object is converted into a watertight surface and normalized within $[-1,1]^3$, ensuring exact coordinate alignment between coarse and fine clouds.

\noindent\textbf{Sharpness-aware sampling.}
To preserve high-frequency features, each point cloud combines uniform surface sampling with curvature-concentrated sampling along with surface normals—faithfully capturing thin structures and sharp boundaries across resolutions.

\noindent\textbf{Community value \& efficiency.}
\noindent \dataset{} provides three key benefits: (i) it eliminates the prohibitive compute and storage overhead of large-scale 3D preprocessing; (ii) it seamlessly interfaces with vecset VAE and DiT for accelerated training; and (iii) it lowers the barrier for high-resolution 3D generation. Detailed dataset statistics are summarized in Table~\ref{tab:sampling_stats}, with visual samples illustrated in Figure~\ref{fig:dataset_samples}.


\begin{table}[t]
\centering
\footnotesize
\setlength{\tabcolsep}{10pt} 
\caption{Geometry generation performance comparison. All models are evaluated after ICP alignment.}
\label{tab:main}
\begin{tabular}{lcc}
\toprule
Method & CD ($\downarrow$) & F-Score$_{0.05}$ ($\uparrow$) \\
\midrule
Hunyuan3D-2.1~\cite{hunyuan3d2025hunyuan3d} & 0.0220 & 0.6389 \\
UltraShape~\cite{jia2025ultrashape}          & 0.0185 & 0.6676 \\
TRELLIS~\cite{xiang2025structured}         & 0.0216 & 0.6400 \\
TRELLIS.2~\cite{xiang2026native}           & 0.0192 & 0.6636 \\
\midrule
\textbf{\method{} (ours)}                   & \textbf{0.0164} & \textbf{0.6886} \\
\bottomrule
\end{tabular}
\vspace{-1em}
\end{table}

\vspace{-0.5em}
\section{Experiments}
\vspace{-0.5em}
\label{sec:exp}


\subsection{Experimental Settings}
\vspace{-0.5em}
We adopt the RoPE-guided vecset DiT architecture from UltraShape. Models are trained with AdamW~\cite{loshchilov2017decoupled} ($\text{lr}=1{\times}10^{-5}$, bfloat16, batch size $64$) for $300\text{k}$ steps on $8$ NVIDIA B200 GPUs. The latent token count is set to $10,240$.

\noindent\textbf{Evaluation Metrics and Benchmark.} Generated shapes are aligned with ground truth via Iterative Closest Point (ICP) before evaluation. We evaluate spatial accuracy and surface fidelity using Chamfer Distance (CD, $\downarrow$) and F-Score ($\uparrow$, with $r=0.05$) on the Toys4K\cite{stojanov2021using}, strictly isolated from training.

\noindent\textbf{Baselines.}
We evaluate our model against SOTA 3D generation baselines: Hunyuan3D-2.1~\cite{hunyuan3d2025hunyuan3d}, TRELLIS~\cite{xiang2025structured}, TRELLIS.2~\cite{xiang2026native}, and UltraShape~\cite{jia2025ultrashape}. All methods are tested under identical ICP alignment with $409,600$ sampled points. 


\vspace{-1em}
\subsection{Geometry Generation}
\vspace{-0.5em}
We evaluate the 3D generation performance of our proposed method against representative state-of-the-art baselines. As quantitatively demonstrated in Table~\ref{tab:main}, our approach consistently outperforms existing methods across all metrics, achieving lower CD alongside higher F-Score, which highlights its superior geometric precision and volumetric completeness. 


Qualitative results are shown in Figure~\ref{fig:qualitative_comparison}. Visual evidence confirms that our model generates high-fidelity 3D meshes with rich details, sharp boundaries, and strict spatial alignment. Unlike prior approaches prone to over-smoothing or topological artifacts, our framework delivers superior surface reconstruction and boundary crispness.

\begin{figure}[t]
\centering
\includegraphics[width=\linewidth]{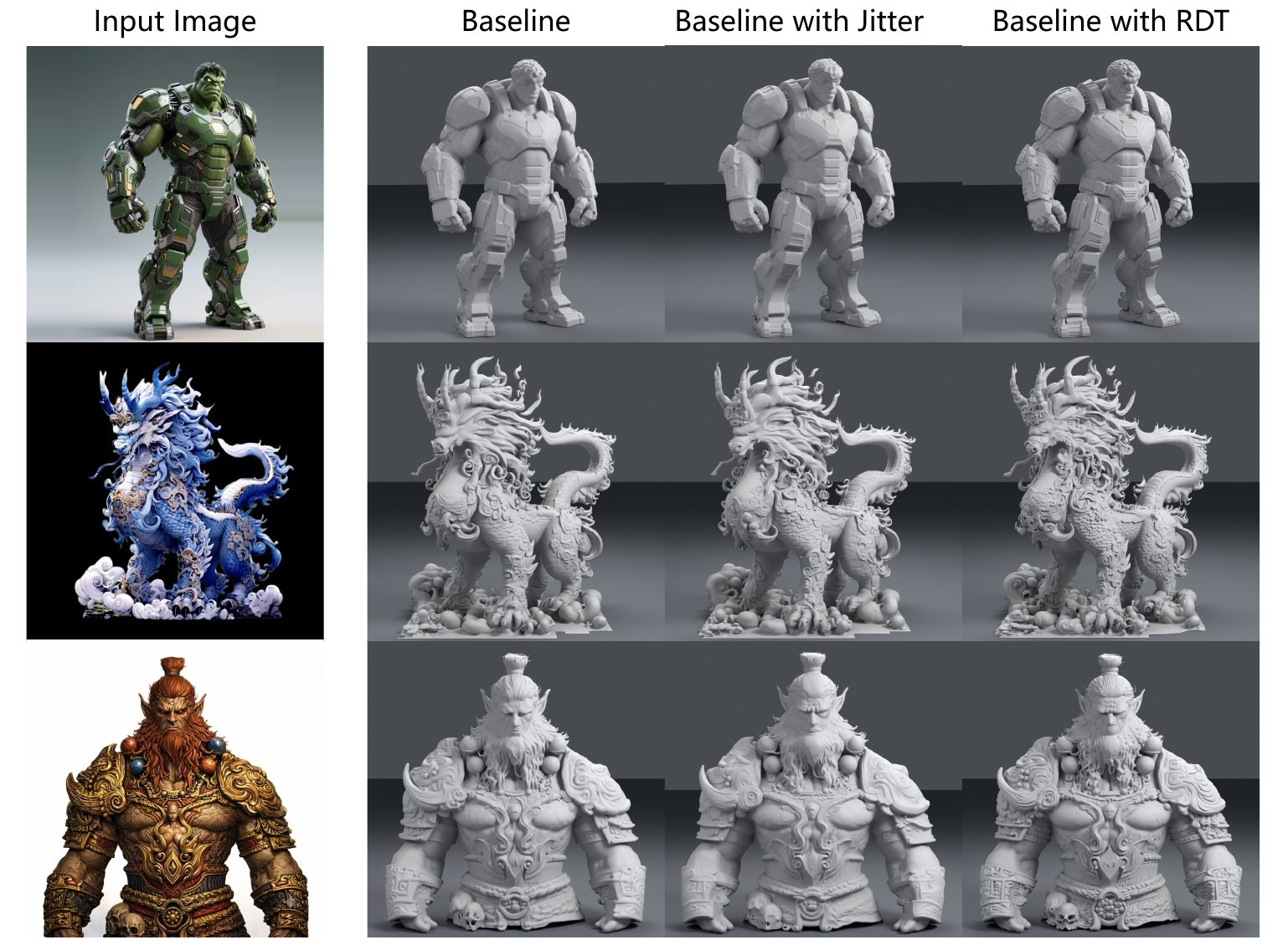}
\caption{\textbf{Qualitative comparison for ablation study.} \rdt{} recovers sharper edges and thinner structures.}
\vspace{-1em}
\label{fig:abl}
\end{figure}




\begin{table}[t]
\centering
\footnotesize
\setlength{\tabcolsep}{3.5pt}
\caption{Ablation study on resolution decoupling and coordinate perturbation.}
\label{tab:abl}
\begin{tabular}{lcccc}
\toprule
Method & $R_g$ & $R_s$ & CD ($\downarrow$) & F-Score ($\uparrow$) \\
\midrule
Baseline & 1536 & 1536 & 0.0184 & 0.6651 \\
Baseline with Jitter & 1536* & 1536 & 0.0187 & 0.6659 \\
\textbf{Baseline with \rdt{}} & \textbf{512} & \textbf{1536} & \textbf{0.0164} & \textbf{0.6886} \\
\bottomrule
\end{tabular}
\vspace{1pt}
\leftline{\scriptsize{\quad * Denotes spatial noise added to RoPE coordinates.}}
\vspace{-3em}
\end{table}

\vspace{-1em}
\subsection{Ablation Studies}
\vspace{-0.5em}

We validate the efficacy of the proposed \rdt{} strategy by comparing three settings under identical model architectures and training hyperparameters with identical coarse guidance (generated by Hunyuan3D-2.1).

\noindent\textbf{Baseline:} uses symmetric $1536$-resolution for both RoPE guidance and geometric supervision ($1536 \!\to\! 1536$).

\noindent\textbf{Baseline with Jitter:} adds random Gaussian spatial noise to RoPE coordinate guidance under symmetric $1536$ supervision.

\noindent\textbf{Baseline with \rdt{}:} explicitly decouples resolutions by anchoring RoPE guidance at $512$ while supervising geometry at $1536$ ($512 \!\to\! 1536$).

\noindent\textbf{Quantitative Evaluation.} 
As shown in Table~\ref{tab:abl}, merely adding coordinate noise yields marginal or inconsistent variations, proving that random perturbation cannot resolve the resolution shift. In contrast, \rdt{} substantially boosts reconstruction performance, confirming that structured resolution decoupling forces robust feature extrapolation from coarse spatial cues. We empirically adopt a $512$ coarse guidance resolution to achieve an optimal balance between quality and computational efficiency.

\noindent\textbf{Qualitative Analysis.} 
Visual comparisons in Figure~\ref{fig:abl} further highlight the structural advantages of \rdt{}. While the baseline produces oversmoothed surfaces and missing details, \rdt{} effectively bridges guidance features and target 3D structures, enabling the recovery of sharp boundaries and fine-grained geometric details.




\vspace{-1em}
\section{Conclusion}
\vspace{-0.5em}
\label{sec:concl}

In this work, we address the training-inference resolution mismatch in RoPE-guided 3D generation via \textit{Resolution-Decoupled Training} (\rdt{}), which anchors RoPE spatial guidance at coarse scale while enforcing ultra-high-resolution supervision to extrapolate fine geometry. Additionally, we introduce \dataset{}, a dataset of $300\text{K}$ watertight point clouds aligned at $512$ and $1536$ resolutions to lower the computational barrier for high-resolution 3D generation.

\bibliographystyle{IEEEbib}
\bibliography{strings,refs}
\end{document}